# Geometric Atlas of the Middle Ear and Paranasal Sinuses for Robotic Applications

Guillaume Michel[1,2], Durgesh Haribhau Salunkhe[2], Philippe Bordure[1] and Damien Chablat[2]

## Abstract
In otolaryngologic surgery, more and more robots are being studied to meet the clinical needs of operating rooms. However, to help design and optimize these robots, the workspace must be precisely defined taking into account patient variability. The aim of this work is to define a geometric atlas of the middle ear and paranasal sinuses for endoscopic robotic applications. Scans of several patients of different ages and sexes were used to determine the average size of these workspaces, which are linked by the similar use of endoscopes in surgery.

## Background

In otologic and sinus surgery, surgeons frequently use endoscopes for minimally invasive purposes. However, the surgeon must hold the endoscope in one hand and can only operate with the other hand [1]. The use of a surgical suction tool is therefore not possible. This constraint makes this type of surgery difficult, and could be improved by the development of medical assistance robots [2].

In order to create a complete set of specifications for the development of assistance robots for otologic and endonasal surgery, the workspace of these robots must be studied in relation to the patient's anatomy. In otology, the external ear and the middle ear are the preferred sites for using the endoscope [3]. The workspace is therefore defined by these two anatomical regions, consisting of (i) the external auditory canal, (ii) the tympanic membrane and (iii) the eardrum.

Sometimes the endoscope is used in the mastoid bone but this clinical situation is less common. In this case, the workspace can be enlarged because the surgeon can drill the bone on demand.

For endonasal surgery, the workspace is defined by the nasal cavity and the paranasal sinuses, including the maxillary, ethmoidal, frontal and sphenoidal sinuses. This workspace is larger and deeper than the ear workspace. We have deliberately defined the sinus workspace up to the side walls of the sinuses, although the endoscope rarely needs to reach the most lateral wall of a sinus. However, this allows the robot to cover all surgical indications, including neoplasms and anterior skull base surgeries [4].

Finally, these two workspaces share several common characteristics: (i) the same scale of dimensions (compared to abdominal or thoracic surgery), (ii) the same endoscopes and (iii) the same surgical techniques. This observation makes it possible to envisage the use of the same robot for otologic and endonasal surgery in order to reduce development costs, training costs and to increase its use in small hospital structures.

The main objective of this paper was to define a geometrical atlas of the middle ear and the paranasal sinuses for endoscopic robotic applications.

The second objective is to know the patient's position on the operating table in order to define the robot's placement constraints in relation to the patient.

The outline of the paper is: (i) a definition of the measurement method, (ii) ear and sinus workspace results, (iii) various positioning of the robot in relation to the workspace, and (iv) discussion with previous works.

## Method

To define these workspaces, we retrospectively selected patients admitted to the Nantes University Hospital for an ear or sinus scanner in 2018. The scans were analysed randomly and anonymously to obtain groups of patients of different sex and age. Patients with anatomical malformations contraindicating endoscopic surgery (such as aplasia) or already operated on with distortions of the bone structure were excluded from the study.

A post-scan study was conducted using Vue PACS v11.3 software from Carestream. For the ear workspace, the following measurements were taken (Figure 1):

- $CAE_{d\_lateral}$: diameter of the external acoustic meatus,
- $CAE_{d\ diameter}$: of the external auditory canal at the tympanic membrane,

[1] Service ORL, CHU de Nantes, France
[2] Laboratoire des Sciences du Numérique de Nantes (LS2N), UMR CNRS 6004, France

**Corresponding author:**
Guillaume Michel, Service ORL, CHU de Nantes, 1 place A.Ricordeau, BP 1005, 44093 Nantes Cedex 01, France
Email: guillaume.michel@chu-nantes.fr

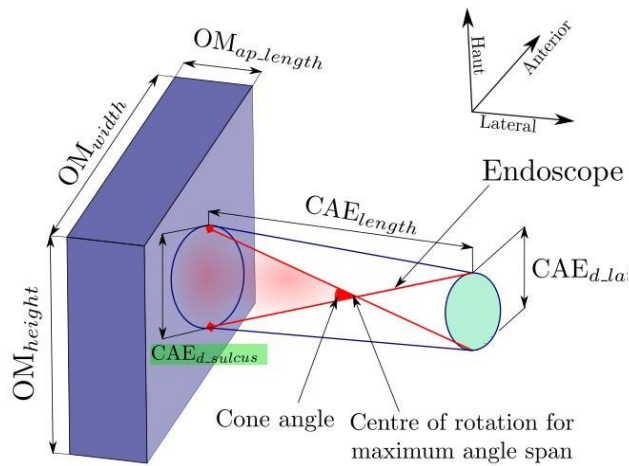

**Figure 1.** Schematic workspace of the external (cylinder) and middle ear.

- $CAE_{lengh}$: length of the external auditory canal in axial section,
- $OM_{height}$: height of the eardrum (from hypotympanic cells to the tegmen tympani),
- $OM_{width}$: length from the retrotympanum to the protympanum,
- $OM_{ap\_length}$: length from the tympanic membrane to the ovale window.

For the sinus workspace, the measurements were (Figure 2):

- *Workspace_depth*: distance between the piriform orifice and the posterior pharyngeal wall,
- *Workspace_width*: distance between the lateral wall of the maxillary sinus and nasal septum on each side,
- *Workspace_height*: distance between the floor of the nasal fossae and the roof of the ethmoid at the level of the naso-frontal canal,
- *Nasal_fossae_width*: distance between the nasal septum and the middle meatus on each side,
- Height of the piriform orifice, which would correspond to the endoscope entrance orifice, the probable site of the remote center of motion.

An anonymised analysis on Excel v.14.5 generated the means and standard deviations. Statistical analysis was performed using SPSS 20 software and Pearson's correlation test.

## Results

### Ear workspace

At the Nantes University Hospital, 36 patients were included from January to March 2018, with an average age of 39 years (2–81). The sex ratio was one to one and 17 right and 19 left ears were analysed.

The average diameter of the external auditory canal was 6.1 mm (3.9–7.3) laterally, and 7.9 mm (6.2–11.1) medially, i.e. at the tympanic membrane level. The average length of the external auditory canal was 26.9 mm (22.5–35.3). The average volume of the external auditory canal was 1.32 cm$^3$.

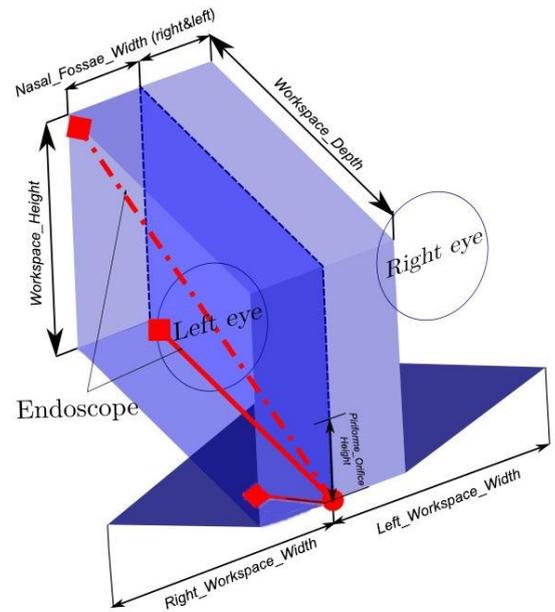

**Figure 2.** Schematic workspace of paranasal sinuses, coronal view: maxillary sinuses (triangles) and vertical workspace from the floor of the nasal cavities to the roof of the ethmoid.

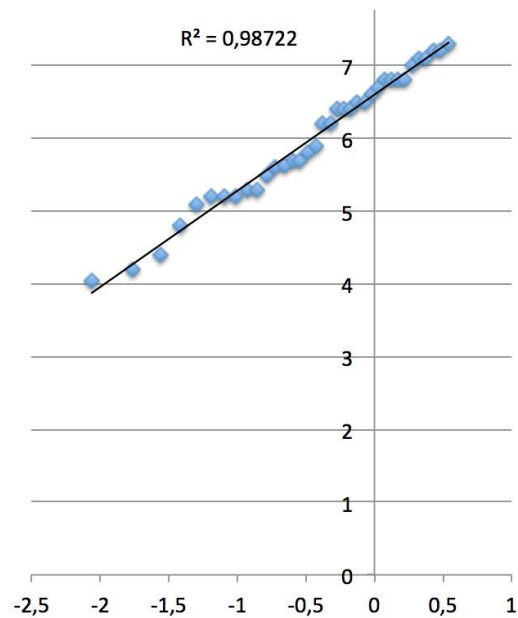

**Figure 3.** Values of the ear workspace, with a standard normal distribution (r$^2$=0.987)

The average height of the eardrum was 16.2 mm (14.1–19.4), while its width was 10.8 mm (7.6–12.3) and its depth was 5.7 mm (3.1–7.2). The average volume of the eardrum was 0.99 cm$^3$.

These values are distributed according to a standard normal distribution ($r^2 > 0.9$), as illustrated in Figure 3. Through percentile analysis, we can estimate that 90% of patients would have an external auditory canal with a diameter between 4.35 and 7.2 mm laterally and between and 10.2 medially, and a length between 22.5 and 34.6 mm. 90% of patients would have an eardrum with a height between 14.3 and 19.1 mm, a width between 8.2 and 13.5 mm, and a depth between 4.3 and 7.3 mm.

Statistical analysis showed that the diameter of the external auditory canal was significantly smaller ~~at entry~~ when age was lower at the meatus ($p = 0.03$), but not at the tympanic membrane level. The length of the external auditory canal was also significantly shorter when the age was lower ($p = 0.013$). No significant differences were found in eardrum measurements by age, nor in workspace measurements for gender or side.

Figure 1 illustrates the ear workspace, with a cylinder depicting the external auditory canal and a rectangular parallelepiped for the eardrum. The red arrows represent the most extreme insertion axes of the endoscope allowed by this workspace, and the theoretical remote center of rotation of the endoscope. We also need the length of the cylinder to define the translation movements of the endoscope.

### Sinus workspace

23 patients have been included, with patients from 11 to 95 years-old. The average depth of the workspace, i.e. the distance between the piriform orifice and the posterior pharyngeal wall, was 77.04 mm (59–94).

The average width of the workspace, i.e. the distance between the lateral wall of the maxillary sinus and nasal septum was 39.26 mm (27–47). Considering the distance without entering the maxillary sinus (distance between the nasal septum and the middle meatus), as usually during surgery, the average distance was 13.74 mm (9–18). This corresponds to the width of the nasal fossae.

The average height of the workspace, i.e. the distance between the floor of the nasal fossae and the roof of the ethmoid, was 55.39 mm (42–67). The average height of the piriform orifice, which delimits the entrance of this workspace, was 29.57 mm (21–36).

These values are distributed according to a standard normal distribution ($r^2 > 0.8$). Through percentile analysis, we can estimate that 90% of patients would have a nasal fossa with a depth between 63.4 and 86.8 mm, a width between 10.1 and 17.5 mm, a height between 47.1 and 66.9 mm. The distance from nasal septum to the lateral wall of the maxillary sinus would be between 31.3 and 44.5 mm. 90% of patients would have a piriform orifice with a height between 24.2 and 34.9 mm.

No significant difference was found depending on age, gender or side. However, only one patient was under 20 years-old. And this particular patient was 11 years-old, so he had all his sinuses formed, although the sphenoid and frontal sinuses were still likely to grow. A series of patients under 10 years old would have shown a significant variation in sinus size compared to adults, as some sinuses would not have been formed yet.

The Figure 2 illustrates the sinus workspace, with two rectangular parallelepipeds for nasal fossae and ethmoidal sinuses up to the posterior pharyngeal wall; and two triangles depicting the two maxillary sinuses.

In red are represented the different positions of the endoscope, with pivot point at the nasal entry. It can be noted that the maximum travel of the endoscope can reach up to 90°, thus emphasizing the importance of having feasible workspace with at least ±45°. But in clinical practice, to access the anterior wall of the maxillary sinus without removing too much bone, the surgeon often uses a 70° endoscope; the travel of the endoscope is thus less than 90°. In some cases, the partition separating the two nasal cavities is removed resulting in an enlarged workspace.

### Position of the robot in relation to the workspace

We have previously defined the different workspaces robotic architecture must adapt. However, robotic architecture must also take into account the variations in the patient's position. During ear or facial surgery, the patient is positioned supine. However, the precise position of the head varies: depending on the type of table and headrest, on the patient's morphology, on the type of surgery, and according to the surgeon's practice.

During otologic surgery, the head is commonly tilted to the opposite side of the operated ear. The mastoid bone must be horizontal: it is the optimal position illustrated in Figure 4, on the left side.

Unfortunately, the patient's anatomy does not always allow the ear to be in this position. For example, an elderly patient sometimes cannot turn his head, because of cervical osteoarthritis for example. Therefore, the mastoid bone cannot be put in the optimal position, and the insertion axis of the endoscope is modified, as illustrated in Figure 4 on the right. Then, the axis of the ear can therefore be tilted by about 53°.

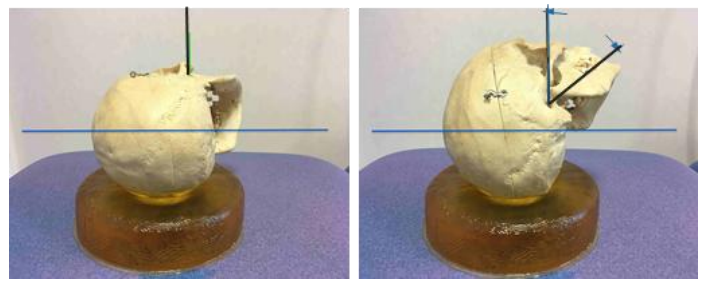

**Figure 4.** Placement of the ear in relation to the operating table in the optimal (right) and unfavourable (left) position

Conversely, during sinus surgery, the head is most often oriented in anterior flexion, in order to have easier access to the antero-superior spaces of the paranasal sinuses, such as the anterior ethmoid and the naso-frontal canal (Figure 5a).

But the patient's morphology can influence the position of the patient's head. Thus, an obese patient, for example, cannot be set correctly in hyperflexion during a sinus surgery. The head will remain horizontal, as represented in Figure 5b. This modified position may be necessary due to medical reasons: in case of cervical arthrodesis for example, flexion of the spine is usually not possible and the patient must remain in a horizontal position.

To objectify these differences in the positioning of the patient's head depending on the situation, we carried out these scenarios with a skull model. We made positions in minimum and maximum situation for ear and sinus surgeries. The choice of the robot architecture is important to allow the robot to adapt to these different situations. Larger the workspace, thanks to high amplitudes of movements, less will the surgeon be constrained by the robot. Indeed, the ear must not be positioned according to the robot, but the robot must adapt to the different positions and morphologies. A choice of architecture based on the middle ear could strongly constrain its field of action to extend the use of the robot to other applications such as sinus surgery or neurosurgery.

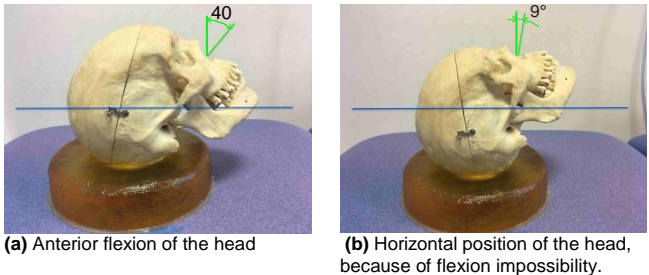

**Figure 5.** Variations in head positions during endonasal surgery. *Axis of the table in blue, insertion axis of the endoscope in green between* $9°$ *and* $40°$.

(a) Anterior flexion of the head

(b) Horizontal position of the head, because of flexion impossibility.

In other words, a robot that is too optimized for the middle ear, for example, could constrain the robot in other operations.

## Discussion

We developed a geometric atlas fitted for robotic ear and sinus endoscopic surgery.

### Ear workspace studies

In the literature, the analyses are most often radiological and concern the external auditory canal, the ossicles or the mastoid. The results are variable, because the definitions of the studied areas change a lot.

Thus, the Parisian team developing Robotol [5] used 12 scanners to measure a workspace corresponding to the surgical speculum, the external auditory canal, and the part of the eardrum visible under microscopy only. They performed a geometric approximation based on these data.

They obtained a 32 × 40 mm cone and a 34 × 16 mm cylinder.

Pacholke [6] calculated the average volume of the middle ear from 15 scans. The middle ear was defined by the tympanic membrane laterally and by the interface between air and temporal bone in all other directions. The volume was estimated around 580 $mm^3$, with a maximum axial dimension of 1570 mm. This result is close to those found by Mas [7], evaluating the same volume around 520–620 $mm^3$ from 18 scans.

The largest study found from 100 scans [8] evaluated the volume of the external auditory canal at 1.4 mL and that of the middle ear at 1.1 mL. This volume decreases significantly in the presence of chronic otitis media. However, this volume, measured in another study based on 91 patients [9], was not significantly different depending on the age or sex of the patient.

Other studies have focused more specifically on the mastoid: Dillon [10], on cadaveric models, and Cros [11], from ten scanners. However, the mastoid is not the preferred working area for endoscopic surgery, and can be enlarged on request by drilling.

Overall, the data available in the literature vary from one study to another, and we did not find geometric measurements of the entire middle ear.

In our study, the average volume of the external auditory canal was 1.32 $cm^3$. This volume is close to the volume found in the literature: 1.4 mL from scanners [8]; between 1.1 and 1.7 mL from tympanometry [12]. The average tympanic volume was 0.99 $cm^3$. This volume is consistent with data found in the literature, between 0.52 $cm^3$ [7] and 1.1 $cm^3$ [8].

Our values are distributed according to a standard normal distribution. Through percentile analysis, it gives us values corresponding to 90% of patients. Maximum values are valuable to calculate the amplitude the robot needs to explore the entire workspace. The minimum values are decisive in knowing when the robot cannot be used. We know that endoscopes have usually a diameter of 2.7 mm, the suction cannula have a diameter between 0.8 to 1.4 mm and the micro-instruments between 1.0 to 2.5 mm (from micro-tip to micro-gripper). Then, for patients with minimum diameter values, it may be difficult to insert all instruments at the lateral end of the external ear canal. A small skin incision should help in this clinical situation. But there must be no difficulty to insert all instruments at the medial extremity of the external auditory canal, even in minimal conditions. According to these data for insertion, no bone drilling should be necessary. However, movements may be more constrained in this particular situation.

### Sinus workspace studies

We found in the literature studies characterizing the workspace for endoscopic surgery. But, as for ear workspace, methodology or definitions are variable, so are the results.

Burgner [13] characterized the workspace to reach the skull base, from 7 scanners (age and gender not specified). This workspace is inadequate for maxillary or frontal sinus surgery, but helpful in terms of depth to the sphenoid and the skull base. The workspace was rectangular and measured 16 × 35 mm. The distance from the nostril to the pituitary gland was 10 cm.

Other authors have studied the workspace from surgical data, and not from scanners. Eichhorn [14] measured the position of the extremity of the endoscope, from 23 endonasal surgeries. The workspace was a cube of 16.59 × 11.38 × 6.30 mm and 1.19 $cm^2$. This workspace was half the size of the one measured in our study.

Another study from Trévillot [15] measured the position of a $30°$ endoscope, from 13 ethmoidectomies in a laboratory. They established boxes containing the pivot point, and depending on the anatomic site to be treated. These boxes were all included into a larger box, where the pivot point should always remain during endonasal surgery. This box was positioned at the nasal entry, and measured from 4.8 to 20.9 mm along the x axis, from 13.8 to 30.9 mm along the y axis and from 7.2 to 34.9 mm along the z axis. These variations were explained by anatomical differences between patients and different surgical techniques. The depth of penetration of the endoscope into the nasal cavity varied between 70 and 100 mm.

These data are consistent with those of Lombard [2], from data recorded with a navigation system during real surgery. The depth of insertion varied from 35 to 112 mm according to the surgery performed.

The average depth, i.e. the distance between the piriform orifice and the posterior pharyngeal wall, was 77.04 mm (59–94) in our study. This dimension corresponds to the insertion depth found in the literature, from 70 to 100 mm for Trévillot [15] and from 35 to 112 mm for Lombard [2].

These studies are highly valuable and gives workspace data to reach the skull base or on the positions of the pivot point. Our study contributes to define delimitation of the entire workspace for an endonasal robotic application.

In contrast to the ear's working space, the piriform orifice is about six times larger than the external auditory canal. There is therefore no problem inserting the endoscope and two auxiliary tools through the nostril at the same time.

## Conclusions

In this study, we defined the ear and sinus workspace measurements in which the endoscope operates. These measurements were taken from a large number of scans made in patients of different ages and sexes, for both locations.

For the ear workspace, the dimensions were consistent with some volumetric data found in the literature [7, 8]. For the sinus workspace, most studies have focused on different sinus volumes, as well as the influence of different pathologies, infectious or malformative, on their size or growth. The maxillary sinus is the most studied sinus [16, 17]. In these studies, it is shown, for example, a decrease in volume with age and loss of maxillary teeth [18]. Other sinuses are also studied in terms of size and anatomical ratios, such as the sphenoid sinus [19]. However, to our knowledge, there is no study that examines the dimensions of the paranasal sinuses as a whole, setting upper and lower limits to define a robotic workspace. These results could help engineers design endoscopic robots adapted to the anatomy of the ear and sinuses.

### Acknowledgements

First author received the grant from "Association Française d'Oto-Neurochirurgie" and the second author receiving financial support from the NExT (Nantes Excellence Trajectory for Health and Engineering) Initiative and the Human Factors for Medical Technologies (FAME) research cluster.

## Declaration of Conflicting Interests

The Authors declares that there is no conflict of interest.

## Supplementary Material

Datasets from Excel for ear and sinus measurements are included as supplementary material.